# Adaptive Load Balancing: A Study in Multi-Agent Learning

**Andrea Schaerf**                                        ASCHAERF@DIS.UNIROMA1.IT
*Dipartimento di Informatica e Sistemistica*
*Università di Roma "La Sapienza", Via Salaria 113, I-00198 Roma, Italy*

**Yoav Shoham**                                           SHOHAM@FLAMINGO.STANFORD.EDU
*Robotics Laboratory, Computer Science Department*
*Stanford University, Stanford, CA 94305, USA*

**Moshe Tennenholtz**                                     MOSHET@IE.TECHNION.AC.IL
*Faculty of Industrial Engineering and Management*
*Technion, Haifa 32000, Israel*

## Abstract

We study the process of multi-agent reinforcement learning in the context of load balancing in a distributed system, without use of either central coordination or explicit communication. We first define a precise framework in which to study adaptive load balancing, important features of which are its stochastic nature and the purely local information available to individual agents. Given this framework, we show illuminating results on the interplay between basic adaptive behavior parameters and their effect on system efficiency. We then investigate the properties of adaptive load balancing in heterogeneous populations, and address the issue of exploration vs. exploitation in that context. Finally, we show that naive use of communication may not improve, and might even harm system efficiency.

## 1. Introduction

This article investigates multi-agent reinforcement learning in the context of a concrete problem of undisputed importance – load balancing. Real life provides us with many examples of emergent, uncoordinated load balancing: traffic on alternative highways tends to even out over time; members of the computer science department tend to use the most powerful of the networked workstations, but eventually find the lower load on other machines more inviting; and so on. We would like to understand the dynamics of such emergent load-balancing systems and apply the lesson to the design of multi-agent systems.

We define a formal yet concrete framework in which to study the issues, called a *multi-agent multi-resource stochastic* system, which involves a set of agents, a set of resources, probabilistically changing resource capacities, probabilistic assignment of new jobs to agents, and probabilistic job sizes. An agent must select a resource for each new job, and the efficiency with which the resource handles the job depends on the capacity of the resource over the lifetime of the job as well as the number of other jobs handled by the resource over that period of time. Our performance measure for the system aims at globally optimizing the resource usage in the system while ensuring fairness (that is, a system shouldn't be made efficient at the expense of any particular agent), two common criteria for load balancing.

How should an agent choose an appropriate resource in order to optimize these measures? Here we make an important assumption, in the spirit of reinforcement learning (Sutton, 1992): The information available to the agent is only its prior experience. In particular, the agent does not necessarily know the past, present, or future capacities of the resources,[1] and is unaware of past, current, or future jobs submitted by the various agents, not even the relevant probability distributions. The goal of each agent is thus to adapt its resource-selection behavior to the behavior of the other agents as well as to the changing capacities of the resources and to the changing load, without explicitly knowing what they are.

We are interested in several basic questions:

- What are good resource-selection rules?

- How does the fact that different agents may use different resource-selection rules affect the system behavior?

- Can communication among agents improve the system efficiency?

In the following sections we show illuminating answers to these questions. The contribution of this paper is therefore twofold. We apply multi-agent reinforcement learning to the domain of adaptive load balancing and we use this basic domain in order to demonstrate basic phenomena in multi-agent reinforcement learning.

The structure of this paper is as follows. In Section 2 we discuss our general setting. The objective of this section is to motivate our study and point to its impact. The formal framework is defined and discussed in Section 3. Section 4 completes the discussion of this framework by introducing the resource selection rule and its parameters, which function as the "control knobs" of the adaptive process. In Section 5 we present experimental results on adaptive behavior within our framework and show how various parameters affect the efficiency of adaptive behavior. The case of heterogeneous populations is investigated in Section 6, and the case of communicating populations is discussed in Section 7. In Section 8 we discuss the impact of our results. In Section 9 we put our work in the perspective of related work. Finally, in Section 10 we conclude with a brief summary.

## 2. The General Setting

This paper applies reinforcement learning to the domain of adaptive load balancing. However, before presenting the model we use and our detailed study, we need to clarify several points about our general setting. In particular, we need to explain the interpretation of reinforcement learning and the interpretation of load balancing we adopt.

Much work has been devoted in the recent years to distributed and adaptive load balancing. One can find related work in the field of distributed computer systems (e.g., Pulidas, Towsley, & Stankovic, 1988; Mirchandaney & Stankovic, 1986; Billard & Pasquale, 1993; Glockner & Pasquale, 1993; Mirchandaney, Towsley, & Stankovic, 1989; Zhou, 1988; Eager, Lazowska, & Zahorjan, 1986), in organization theory and management science (e.g., Malone, 1987), and in distributed AI (e.g., Bond & Gasser, 1988). Although some motivations of the above-mentioned lines of research are similar, the settings discussed have some essential differences.

---

1. In many applications the capacities of the resources are known, at least to some extent. This point will be discussed later. Basically, in this paper we wish to investigate how far one can go using only purely local feedback and without the use of any global information (Kaelbling, 1993; Sutton, 1992).

Work on distributed computer systems adopts the view of a set of computers each of which controls certain resources, has an autonomous decision-making capability, and jobs arrive to it in a dynamic fashion. The decision-making agents of the different computers (also called nodes) try to share the system load and coordinate their activities by means of communication. The actual action to be performed, based on the information received from other computers, may be controlled in various ways. One of the ways adopted to control the related decisions is through learning automata (Narendra & Thathachar, 1989).

In the above-mentioned work each agent is associated with a set of resources, where both the agent and the related resources are associated with a node in the distributed system. Much work in management science and in distributed AI adopts a somewhat complementary view. In difference to classical work in distributed operating systems, an agent is not associated with a set of resources that it controls. The agents are autonomous entities which negotiate among themselves (Zlotkin & Rosenschein, 1993; Kraus & Wilkenfeld, 1991) on the use of shared resources. Alternatively, the agents (called managers in this case) may negotiate the task to be executed with the processors which may execute it (Malone, 1987).

The model we adopt has the flavor of models used in distributed AI and organization theory. We assume a strict separation between agents and resources. Jobs arrive to agents who make decisions about where to execute them. The resources are passive (i.e., do not make decisions). A typical example of such a setting in a computerized framework is a set of PCs, each of which is controlled by a different user and submits jobs to be executed on one of several workstations. The workstations are assumed to be independent of each other and shared among all the users. The above example is a real-life situation which motivated our study and the terminology we adopt is taken from such a framework. However, there are other real-life situations related to our model in areas different from classical distributed computer systems.

A canonical problem related to our model is the following one (Arthur, 1994): An agent, embedded in a multi-agent system, has to select among a set of bars (or a set of restaurants). Each agent makes an autonomous decision but the performance of the bar (and therefore of the agents that use it) is a function of its capacity and of the number of agents that use it. The decision of going to a bar is a stochastic process but the decision of which bar to use is an autonomous decision of the respective agent. A similar situation arises when a product manager decides which processor to use in order to perform a particular task. The model we present in Section 3 is a general model where such situations can be investigated. In these situations a job arrives to an agent (rather than to a node consisting of particular resources) who decides upon the resource (e.g., restaurant) where his job should be executed; there is a-priori no association between agents and resources.

We now discuss the way the agents behave in such a framework. The common theme among the above-mentioned lines of research is that load-balancing is achieved by means of communication among active agents or active resources (through the related decision-making agents). In our study we adopt a complementary view. We consider agents who act in a *purely local* fashion, based on *purely local information* as described in the recent reinforcement learning literature. As we mentioned, learning automata were used in the field of distributed computer systems in order to perform adaptive load balancing. Nevertheless, the related learning procedures rely heavily on communication among agents (or among decision-making agents of autonomous computers). Our work applies recent work on reinforcement learning in AI where the information the agent gets is purely local. Hence, an agent will know how efficient the service in a restaurant has been only by choosing it as a place to eat. We don't assume that agents may be informed by other agents about the load in other restaurants or that the restaurants will announce their current load. This makes our work strictly different from other work applying reinforcement learning to adaptive load balancing.

The above features make our model and study both basic and general. Moreover, the above discussion raises the question of whether reinforcement learning (based on purely local information and feedback) can guarantee useful load balancing. The combination of the model we use and our perspective on reinforcement learning makes our contribution novel. Nevertheless, as we mentioned above (and as we discuss in Section 9) the model we use is not original to us and captures many known problems and situations in distributed load balancing. We apply reinforcement learning, as discussed in the recent AI literature, to that model and investigate the properties of the related process.

## 3. The Multi-Agent Multi-Resource Stochastic System

In this section we define the concrete framework in which we study dynamic load balancing. The model we present captures adaptive load balancing in the general setting mentioned in Section 2. We restrict the discussion to discrete, synchronous systems (and thus the definition below will refer to $\mathcal{N}$, the natural numbers); similar definitions are possible in the continuous case. We concentrate on the case where a job can be executed using any of the resources. Although somewhat restricting, this is a common practice in much work in distributed systems (Mirchandaney & Stankovic, 1986).

**Definition 3.1** *A multi-agent multi-resource stochastic system is a 6-tuple* $\langle \mathcal{A}, \mathcal{R}, \mathcal{P}, \mathcal{D}, \mathcal{C}, SR \rangle$, *where* $\mathcal{A} = \{a_1, \ldots, a_N\}$ *is a set of agents,* $\mathcal{R} = \{r_1, \ldots, r_M\}$ *is a set of resources,* $\mathcal{P} : \mathcal{A} \times \mathcal{N} \to [0,1]$ *is a job submission function,* $\mathcal{D} : \mathcal{A} \times \mathcal{N} \to \Re$ *is a probabilistic job size function,* $\mathcal{C} : \mathcal{R} \times \mathcal{N} \to \Re$ *is a probabilistic capacity function, and* $SR$ *is a resource-selection rule.*

The intuitive interpretation of the system is as follows. Each of the resources has a certain capacity, which is a real number; this capacity changes over time, as determined by the function $\mathcal{C}$. At each time point each agent is either idle or engaged. If it is idle, it may submit a new job with probability given by $\mathcal{P}$. Each job has a certain size which is also a real number. The size of any submitted job is determined by the function $\mathcal{D}$. (We will use the unit *token* where referring to job sizes and resource capacities, but we do not mean that tokens come only in integer quantities.) For each new job the agent selects one of the resources. This choice is made according to the rule SR; since there is much to say about this rule, we discuss it separately in the next section.

In our model, any job may run on any resource. Furthermore, there is no limit on the number of jobs served simultaneously by a given resource (and thus no queuing occurs). However, the quality of the service provided by a resource at a given time deteriorates with the number of agents using it at that time. Specifically, at every time point the resource distributes its current capacity (i.e., its tokens) equally among the jobs being served by it. The size of each job is reduced by this amount and, if it drops to (or below) zero, the job is completed, the agent is notified of this, and becomes idle again. Thus, the execution time of a job $j$ depends on its size, on the capacity over time of the resource processing it, and on the number of other agents using that resource during the execution of $j$.

Our measure of the system's performance will be twofold: We aim to minimize time-per-token, averaged over all jobs, as well as to minimize the standard deviation of this random variable. Minimizing both quantities will ensure overall system efficiency as well as fairness. The question is which selection rules yield efficient behavior; so we turn next to the definition of these rules.

## 4. Adaptive Resource-Selection Rules

The rule by which agents select a resource for a new job, the *selection rule* (SR), is the heart of our adaptive scheme and the topic of this section. Throughout this section and the following one we make an assumption of homogeneity. Namely, we assume that all the agents use the same SR. Notice that although the system is homogeneous, each agent will act based only on its local information. In Sections 6 and 7 we relax the homogeneity assumption and discuss heterogeneous and communicating populations.

As we have already emphasized, among all possible adaptive SRs we are interested in purely local SRs, ones that have access only to the experience of the particular agent. In our setting this experience consists of results of previous job submissions; for each job submitted by the agent and already completed, the agent knows the name $r$ of the resource used, the point in time, $t_{start}$, the job started, the point in time, $t_{stop}$, the job was finished, and the job size $S$. Therefore, the input to the SR is, in principle, a list of elements in the form $(r, t_{start}, t_{stop}, S)$. Notice that this type of input captures the general type of systems we are interested in. Basically, we wish to assume as little as possible about the information available to an agent in order to capture real loosely-coupled systems where more global information is unavailable.

Whenever agent $i$ selects a resource for its job execution, $i$ may get its feedback after non-negligible time, where this feedback may depend on decisions made by other agents before and after agent $i$'s decision. This forces the agent to rely on a non-trivial portion of its history and makes the problem much harder.

There are uncountably many possible adaptive SRs and our aim is not to gain exhaustive understanding of them. Rather, we have experimented with a family of intuitive and relatively simple SRs and have compared them with some non-adaptive ones. The motivation for choosing our particular family of SRs is partially due to observations made by cognitive psychologists on how people tend to behave in multi-agent stochastic and recurrent situations. In principle, our set of SRs captures the two most robust aspects of these observations: "The law of effect" (Thronkide, 1898) and the "Power law of practice" (Blackburn, 1936). In our family of rules, called $\Omega$, which partially resembles the learning rules discussed in the learning automata literature (Narendra & Thathachar, 1989), and partially resembles the interval estimation algorithm (Kaelbling, 1993), agents do not maintain complete history of their experience. Instead, each agent, $A$, condenses this history into a vector, called the *efficiency estimator*, and denoted by $ee_A$. The length of this vector is the number of resources, and the $i$'th entry in the vector represents the agent's evaluation of the current efficiency of resource $i$ (specifically, $ee_A(R)$ is a positive real number). This vector can be seen as the state of a learning automaton. In addition to $ee_A$, agent $A$ keeps a vector $jd_A$, which stores the number of completed jobs which were submitted by agent $A$ to each of the resources, since the beginning of time. Thus, within $\Omega$, we need only specify two elements:

1. How agent $A$ updates $ee_A$ when a job is completed

2. How agent $A$ selects a resource for a new job, given $ee_A$ and $jd_A$

Loosely speaking, $ee_A$ will be maintained as a weighted sum of the new feedback and the previous value of $ee_A$, and the resource selected will most probably be the one with highest $ee_A$ entry except that with low probability some other resource will be chosen. These two steps are explained more precisely in the following two subsections.

## 4.1 Updating the Efficiency Estimator

We take the function updating $ee_A$ to be

$$ee_A(R) := WT + (1 - W)ee_A(R)$$

where $T$ represents the time-per-token of the newly completed job and is computed from the feedback $(R, t_{start}, t_{stop}, S)$ in the following way:[2]

$$T = (t_{stop} - t_{start})/S$$

We take $W$ to be a real value in the interval $[0, 1]$, whose actual value depends on $jd_A(R)$. This means that we take a weighted average between the new feedback value and the old value of the efficiency estimator, where $W$ determines the weights given to these pieces of information. The value of $W$ is obtained from the following function:

$$W = w + (1 - w)/jd_A(R)$$

In the above formula $w$ is a real-valued constant. The term $(1 - w)/jd_A(R)$ is a correcting factor, which has a major effect only when $jd_A(R)$ is low; when $jd_A(R)$ increases, reaching a value of several hundreds, this term becomes negligible with respect to $w$.

## 4.2 Selecting the Resource

The second ingredient of adaptive SRs in $\Omega$ is a function $pd_A$ selecting the resource for a new job based on $ee_A$ and $jd_A$. This function is probabilistic. We first define the following function

$$pd'_A(R) := \begin{cases} ee_A(R)^{-n} & \text{if } jd_A(R) > 0 \\ E[ee_A]^{-n} & \text{if } jd_A(R) = 0 \end{cases}$$

---

2. Using parallel processing terminology, $T$ can be viewed as a *stretch factor*, which quantifies the stretching of a program's processing time due to multiprogramming (Ferrari, Serazzi, & Zeigner, 1983).

where $n$ is a positive real-valued parameter and $E[ee_A]$ represents the average of the values of $ee_A(R)$ over all resources satisfying $jd_A(R) > 0$. To turn this into a probability function, we define the $pd_A$ as the normalized version of $pd'_A$:

$$pd_A(R) := pd'_A(R)/\sigma$$

where $\sigma = \Sigma_R pd'_A(R)$ is a normalization factor.[3]

The function $pd_A$ clearly biases the selection towards resources that have performed well in the past. The strength of the bias depends on $n$; the larger the value of $n$, the stronger the bias. In extreme cases, where the value of $n$ is very high (e.g., $\geq 20$), the agent will always choose the resource with the best record. This strategy of "always choosing the best", although perhaps intuitively appealing, is in general not a good one; it does not allow the agent to exploit improvements in the capacity or load on other resources. We discuss this SR in the following subsection, and expand on the issue of exploration versus exploitation in Sections 6 and 7.

To summarize, we have defined a general setting in which to investigate emergent load balancing. In particular, we have defined a family of adaptive resource-selection rules, parameterized by a pair $(w, n)$. These parameters serve as knobs with which we tune the system so as to optimize its performance. In the next section we turn to experimental results obtained with this system.

## 4.3 The Best Choice SR (BCSR)

The Best Choice SR (BCSR) is a learning rule that assumes a high value of $n$, i.e, which always chooses the best resource in a given point. We will assume $w$ is fixed to a given value while discussing BCSR. In our previous work (Shoham & Tennenholtz, 1992, 1994), we showed that learning rules that strongly resemble BCSR are useful for several natural multi-agent learning settings. This suggests that we need to carefully study it in the case of adaptive load balancing. As we will demonstrate, BCSR is not always useful in the load balancing setting.

The difference between BCSR and a learning rule where the value of $n$ is low, is that in the latter case the agent gives relatively high probability for the selection of a resource that didn't give the best results in the past. In that case the agent might be able to notice that the behavior of one of the resources has been improved due to changes in the system.

Note that the exploration of "non-best" resources is crucial when the dynamics of the system includes changes in the capacities of the resources. In such cases, the agent could not take advantage of possible increases in the capacity of resources if it uses the BCSR. One might wonder, however, whether in cases where the main dynamic changes of the system stem from load changes, relying on BCSR is sufficient. If the latter is true, we will be able to ignore the parameter $n$ and to concentrate only on the BCSR, in systems where the capacity of resources is fixed. In order to clarify this point, we consider the following example.

---

3. If for all $R$ we have $jd_A(R) = 0$, (i.e., if the agent is going to submit its very first job), then we assume the agent chooses a resource randomly (with a uniform probability distribution).

Suppose there are only two resources, $R_1$ and $R_2$, whose respective (fixed) capacities, $c_{R_1}$ and $c_{R_2}$, satisfy the equality $c_{R_1} = 2c_{R_2}$. Assume now that the load of the system varies between a certain low value and a certain high one.

If the system's load is low and the agents adopt BCSR, then the system will evolve in a way where almost all of the agents would be preferring $R_1$ to $R_2$. This is due to the fact that, in the case of low load, there are only few overlaps of jobs, hence $R_1$ is much more efficient. On the other hand, when the system's load is high, $R_1$ could be very busy and some of the agents would then prefer $R_2$, since the performance obtained using the less crowded resource $R_2$ could be better than the one obtained using the overly crowded resource $R_1$. In the extreme case of a very high load, we expect the agents to use $R_2$ one third of the time.

Assume now that the load of the system starts from a low level, then increases to a high value, and then decreases to reach its original value. When the load increases, the agents, that were mostly using $R_1$, will start observing that $R_1$'s performance is becoming worse and, therefore, following the BCSR they will start using $R_2$ too. Now, when the load decreases, the agents which were using $R_2$ will observe an improvement in the performance of $R_2$, but the value they have stored for $R_1$ (i.e., $ee_A(1)$), will still reflect the previous situation. Hence, the agents will keep on using $R_2$, ignoring the possibility of obtaining much better results if they moved back to $R_1$. In this situation, the randomized selection makes the agents able to use $R_1$ (with a certain probability) and therefore some of them may discover that the performance of $R_1$ is better than that of $R_2$ and switch back to $R_1$. This will improve the system's efficiency in a significant manner.

The above example shows that the BCSR is, in the general case, not a good choice. This is in general true when the value of $n$ is too high.

In the above discussion we have assumed that the changes in the load are unforeseen. If we are able to predict the changes in the load, the agents can simply use the BCSR while the load is fixed and then use a low value of $n$ during the changes. In our case, instead, without even realizing that the system has changed in some way, the agents would need to (and, as we will see, would be able to) adapt to dynamic changes as well as to each other.

## 5. Experimental Results

In this section we compare SRs in $\Omega$ to each another, as well as to some non-adaptive, benchmark selection rules.

The non-adaptive SRs we consider in this paper are those in which the agents partition themselves according to the capacities and the load of the system in a fixed predetermined manner and each agent uses always the same resource. Later in the paper, a SR of this kind is identified by a *configuration* vector, which specifies, for each resource, how many agents use it. When we test our adaptive SRs, we compare the performance against the non-adaptive SRs that perform best on the particular problem. This creates a highly competitive set of benchmarks for our adaptive SRs.

In addition, we compare our adaptive SRs to the *load-querying* SR which is defined as follows: Each agent, when it has a new job, asks all the resources how busy they are and always chooses the less crowded one.

## 5.1 An Experimental Setting

We now introduce a particular experimental setting, in which many of the results described below were obtained. We present it in order to be concrete about the experiments; however, the qualitative results of our experiments were observed in a variety of other experimental settings.

One motivation of our particular setting stems from the PCs and workstations problem mentioned in Section 2. For example, part of our study is related to a set of computers located at a single site. These computers have relatively high load with some peak hours during the day and a low load at night (i.e., the chances a user of a PC submits a job is higher during the day time of the week days than at night and on weekend). Another part of our study is related to a set of computers split all around the world, where the load has quite random structure (i.e., due to difference in time zones, users may use PCs in unpredictable hours).

Another motivation of our particular setting stems from the restaurant problem mentioned in Section 2 (for discussion on the related "bar problem" see Arthur, 1994). For example, we can consider a set of snack bars located at an industrial park. These snack bars have relatively high loads with some peak hours during the day and low load at night (i.e., the chances an employee will choose to go to a snack-bar is higher during the day because there are more employees present during the day). Conversely, we can assume a set of bars near an airport where the load has quite random structure (i.e., the airport employees may like to use these snack-bars in quite unpredicted hours).

Although these are particular real-situations, we would like to emphasize the general motivation of our study and the fact that the related phenomena have been observed in various different settings.

We take $N$, the number of agents, to be 100, and $M$, the number of resources, to be 5. In the first set of experiments we take the capacities of the resources to be fixed. In particular, we take them to be $c_1 = 40$, $c_2 = 20$, $c_3 = 20$, $c_4 = 10$, $c_5 = 10$. We assume that all agents have the same probability of submitting a new job. We also assume that all agents have the same distribution over the size of jobs they submit; specifically, we assume it to be a uniform distribution over the integers in the range [50,150].

For ease of exposition, we will assume that each point in time corresponds to a second, and we consequently count the time in minutes, hours, days, and weeks. The hour is our main point of reference; we assume, for simplicity, that the changes in the system (i.e., load change and capacity change) happen only at the beginning of a new hour. The probability of submitting a job at each second, which corresponds to the load of the system, can vary over time; this is the crucial factor to which the agents must adapt. Note that agents can submit jobs at any second, but the probability of such submission may change. In particular we concentrate on three different values of this quantity, called $L_{lo}, L_{hi}$ and $L_{peak}$, and we assume that the system load switches between those values. The actual values of $L_{lo}, L_{hi}$ and $L_{peak}$ in the following quantitative results are 0.1%, 0.3% and 1%, which roughly correspond to each agent submitting 3.6, 10.8, and 36 jobs per hour (per agent) respectively.

| load | configuration | time-per-token |
|------|---------------|----------------|
| $L_{lo}$ | $\{100, 0, 0, 0, 0\}$ | 38.935 |
| $L_{hi}$ | $\{66, 16, 16, 1, 1\}$ | 60.768 |
| $L_{peak}$ | $\{40, 20, 20, 10, 10\}$ | 196.908 |

Figure 1: Best non-adaptive SRs for fixed load

In the following, when measuring success, we will refer only to the average time-per-token.[4] However, the adaptive SRs that give the best average time-per-token were also found to be fair.

## 5.2 Fixed Load

We start with the case in which the load is fixed. This case is not the most interesting for adaptive behavior; however, a satisfactory SR should show reasonably efficient behavior in that basic case, in order to be useful when the system stabilizes.

We start by showing the behavior of non-adaptive benchmark SRs in the case of fixed load.[5] Figure 1 shows those that give the best results, for each of the three loads.

As we can see, there is a big difference between the three loads mentioned above. When the load is particularly high, the agents should scatter around all the resources at a rate proportional to their capacities; when the load is low they should all use the best resource. Given the above, it is easy to see that an adaptive SR can be effective only if it enables moving quickly from one configuration to the other.

In a static setting such as this, we can expect the best non-adaptive SRs to perform better than adaptive ones, since the information gained by the exploration of the adaptive SRs can be built-in in the non-adaptive ones. The experimental results confirm this intuition, as shown in Figure 2 for $L_{hi}$. The figure shows the performance obtained by the population when the value of $n$ varies between 2 to 10 and for three values of $w$: 0.1, 0.3, and 0.5. Note that for the values of $(n, w)$ that are good choices in the dynamic cases (see later in the paper, values in the intervals $[3, 5]$ and $[0.1, 0.5]$, respectively), the deterioration in the performance of the adaptive SRs with respect to the non-adaptive ones is small. This is an encouraging result, since adaptive SRs are meant to be particularly suitable for dynamic systems. In the following subsections we see that indeed they are.

## 5.3 Changing Load

We now begin to explore more dynamic settings. Here we consider the case in which the load on the system (that is, the probability of agents submitting a job at any time) changes over time. In this paper we present two dynamic settings: One in which the load changes according to a fixed pattern with only a few random perturbations and another in which the load varies in some random fashion. Specifically, in the first case we fix the load to be $L_{hi}$ for ten consecutive hours, for five days a week, with two randomly chosen hours in which it is $L_{peak}$, and to be $L_{lo}$ for the rest of the week. In the second case, we fix the number of hours in a week for each load as in the first case, and we distribute them completely randomly in a week.

---

4. In the data shown later we refer, for convenience, to the time for 1000 tokens.
5. The non-adaptive SRs are human-designed SRs that are used as benchmarks; they assume knowledge of the load and capacity, which is not available for the adaptive SRs we design.

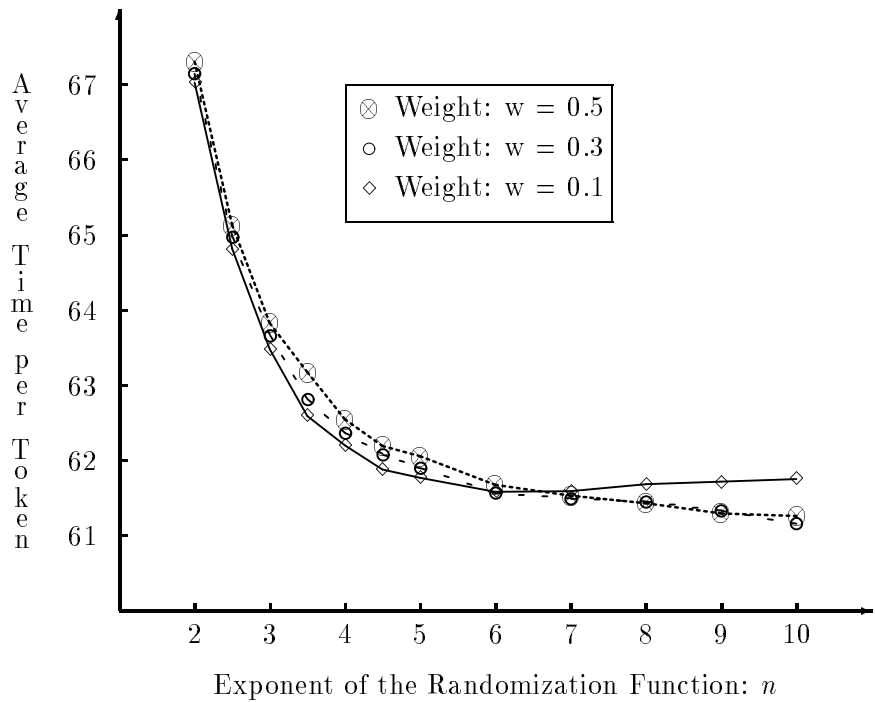

Figure 2: Performance of the adaptive Selection Rules for fixed load

The results obtained for the two cases are similar. Figure 3 shows the results obtained by the adaptive SRs in the case of random load. The best non-adaptive deterministic SR gives the time-per-token value of 69.201 obtained with the configuration (partition of agents) $\{52, 22, 22, 2, 2\}$; the adaptive SRs are superior. The load-querying SR instead gets the time-per-token value of 48.116, which is obviously better, but is not so far from the performances of the adaptive SRs.

We also observe the following phenomenon: Given a fixed $n$ (resp. a fixed $w$) the average time-per-token is non-monotonic in $w$ (resp. in $n$). This phenomenon is strongly related to the issue of exploration versus exploitation mentioned before and to phenomena observed in the study of Q-learning (Watkins, 1989).

We also notice how the two parameters $n$ and $w$ interplay. In fact, for each value of $w$ the minimum of the time per token value is obtained with a different value of $n$. More precisely, the higher $w$ is the lower $n$ must be in order to obtain the best results. This means that, in order to obtain high performance, highly exploratory activity (low $n$) should be matched with giving greater weight to the more recent experience (high $w$). This "parameter matching" can be intuitively explained in the following qualitative way: The exploration activity pays because it allows the agent to detect changes in the system. However, it is more effective if, when a change is detected, it can significantly affect the efficiency estimator (i.e., if $w$ is high). Otherwise, the cost of the exploration activity is greater than its gain.

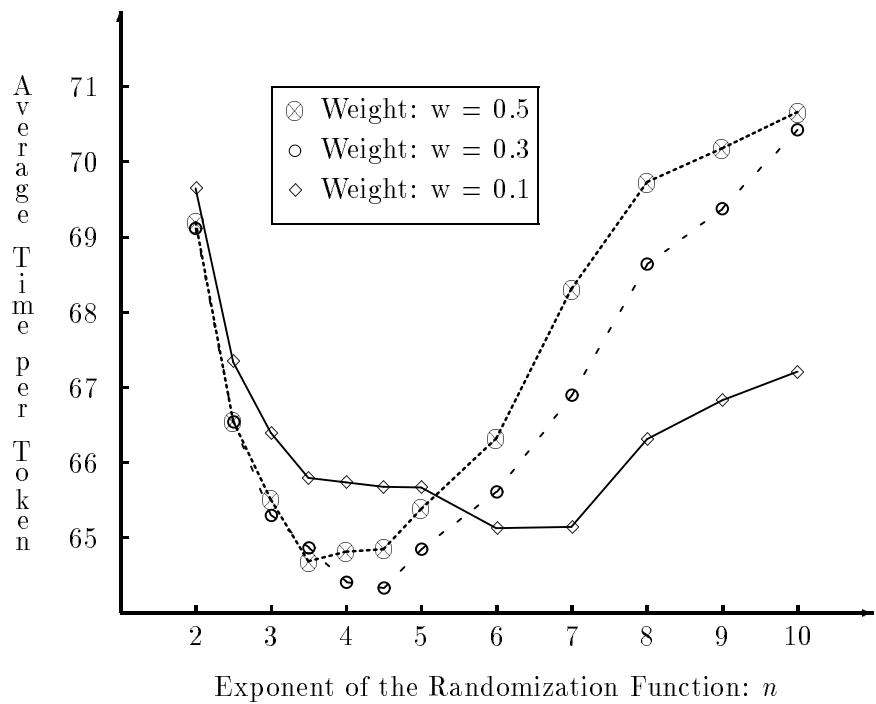

Figure 3: Performance of the adaptive Selection Rules for random load

## 5.4 Changing Capacities

We now consider the case in which the capacity of the resources can vary over time. In particular, we will demonstrate our results in the case of the previously mentioned setting. We will assume the capacities rotate randomly among the resources and, in five consecutive days, each resource gets the capacity of 40 for one day, 20 for 2 days, and 10 for the other 2 days.[6] The load also varies randomly.

The results of this experiment are shown in Figure 4. The best non-adaptive SR in this case gives the time-per-token value of 118.561 obtained with the configuration $\{20, 20, 20, 20, 20\}$.[7] The adaptive SRs give much better results, which are only slightly worse than in the case of fixed capacities. The phenomena we mentioned before are visible in this case too. See for example how a weight of 0.1 mismatches with the low values of $n$.

---

6. Usually the capacities will change in a less dramatic fashion. We use the above-mentioned setting in order to demonstrate the applicability of our approach under severe conditions.

7. The load-querying SR gives the same results as in the case of fixed capacities, because such SR is obviously not influenced by the change.

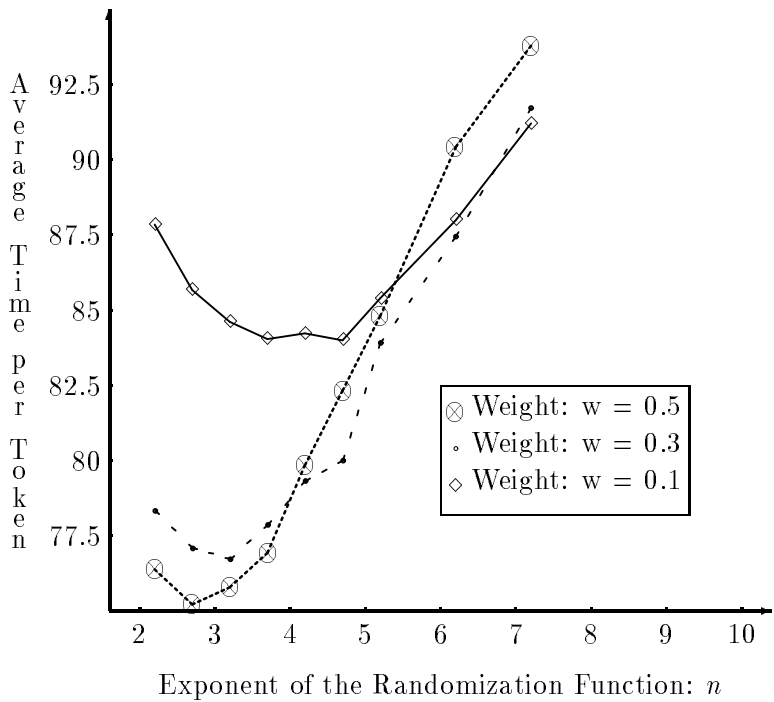

Figure 4: Performance of the adaptive Selection Rules for changing capacities

## 6. Heterogeneous Populations

Throughout the previous section we have assumed that all the agents use the same SR, i.e. Homogeneity Assumption. Such assumption models the situation in which there is a sort of centralized *off-line* controller which, in the beginning, tells the agents how to behave and then leaves the agents to make their own decisions.

The situation described above is very different from having an *on-line* centralized controller which makes every decision. However, we would like now to move even further from that and investigate the situation in which each agent is able to make its own decision about which strategy to use and, maybe, adjust it over time.

As a step toward the study of systems of this kind, we drop the Homogeneity Assumption and consider the situation in which part of the population uses one SR and the other part uses a second one.

In the first set of experiments, we consider the setting discussed in Subsection 5.1 and we confront one with the other, two populations (called 1 and 2) of the same size (50 agents each). Each population uses a different SR in $\Omega$. The SR of population $i$ (for $i = 1, 2$) will be determined by the pair of parameters $(w_i, n_i)$. The measure of success of population $i$ will be defined as the average time-per-token of its members, and will be denoted by $T_i$.

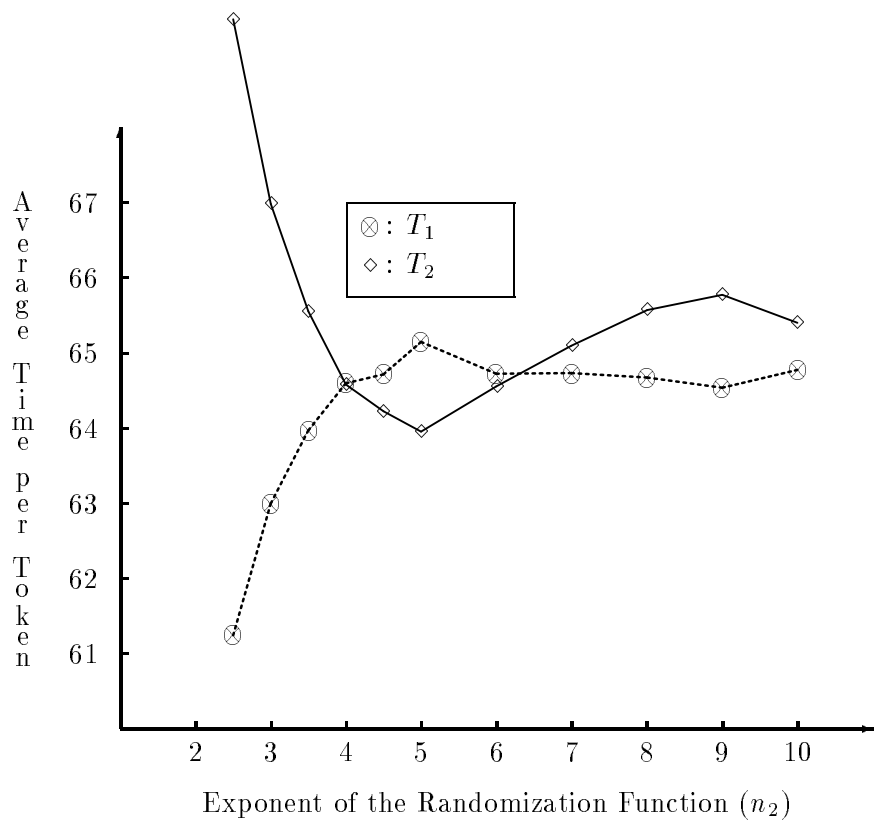

Figure 5: Performance of 2 populations of 50 agents with $n_1 = 4$ and $w_1 = w_2 = 0.3$

Figure 5 shows the result obtained for $w_1 = w_2 = 0.3$, and $n_1 = 4$, and for different values of $n_2$, in the case of randomly varying load.

Our results expose the following phenomenon: The two populations obtain different outcomes from the ones they obtain in the homogeneous case. More specifically, for $4 \leq n_2 \leq 6$ , the results obtained by the agents which use $n_2$ are generally better than the results obtained by the ones which use $n_1$, despite the fact that an homogeneous population which uses $n_1$ gets better results than an homogeneous population which uses $n_2$.

The phenomenon described above has the following intuitive explanation. For $n_2$ in the above-mentioned range, the population which uses $n_2$ is less "exploring" (i.e., more "exploiting") than the other one, and when it is left on its own it might not be able to adapt to the changes in a satisfactory manner. However, when it is joined with the other population, it gets the advantages of the experimental activity of agents in that population, without paying for it. In fact, the more exploring agents, in trying to unload the most crowded resources, make a service to the other agents as well.

It is worth observing in Figure 5 that when $n_2$ is low (e.g., $n_2 \leq 3$) the agents that use $n_2$ take the role of explorers and lose a lot, while the agents that use $n_1$ gain from that situation. Conversely, for high values of $n_2$ (e.g., $n_2 \geq 7$) the performances of the exploiters, which use $n_2$, deteriorate. This means that if the exploiters are too static, then they hinder each other, and the explorers can take advantage of it.

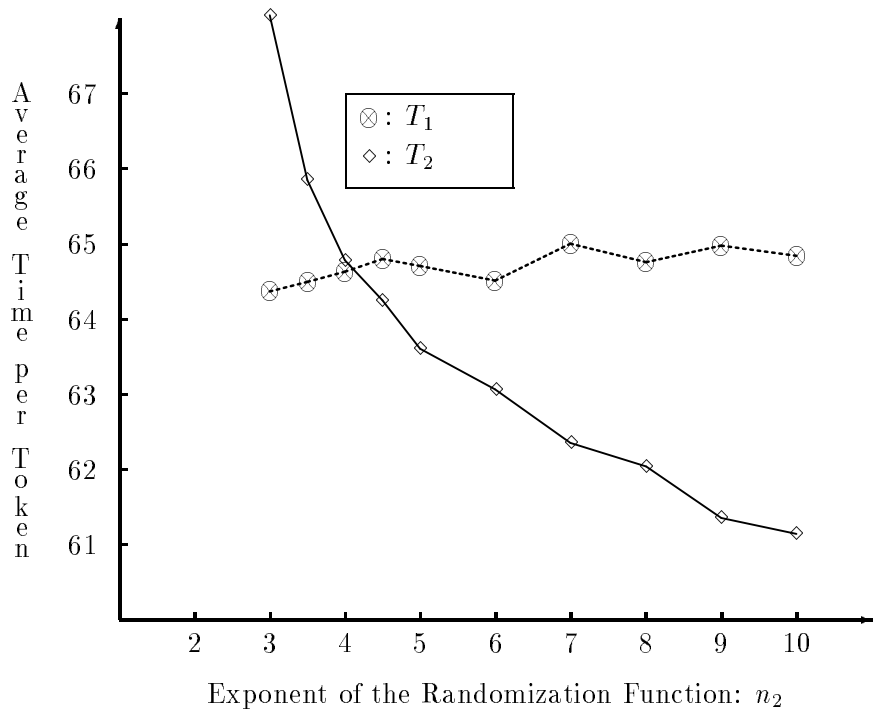

Figure 6: Performance of 2 populations of 90/10 agents with $n_1 = 4$ and $w_1 = w_2 = 0.3$

For a better understanding of the phenomena involved, we have experimented with an asymmetric population, composed of one large group and one small one, instead of two groups of similar size. Figure 6 shows the results obtained using a setting similar to the one above, but where population 1 is composed of 90 members while population 2 consists of only 10 members. In this case, for every value of $n_2 \geq 4$, the exploiters do better than the explorers. The experiments also show that in this case, the higher $n_2$ is the better $T_2$ is, i.e. the more the exploiters exploit, the more they gain.

The above results suggest that a single agent gets the best results for itself by being non-cooperative and always adopting the resource with the best performance (i.e., use BCSR), given that the rest of the agents use an adaptive (i.e., cooperative) SR. However, if all of the agents are non-cooperative then all of them will lose.[8] In conclusion, the selfish interest of an agent does not match with the interest of the population. This is contrary to results obtained in other basic contexts of multi-agent learning (Shoham & Tennenholtz, 1992).

What we have shown is how, for a fixed value of $w$, coexisting populations adopting different values of $n$ interact. Similar results are obtained when we fix the value of $n$ and use two different values for $w$. In such cases, the agents adopting the lower value of $w$ are in general the winners, as shown in Figure 7 for $n_1 = n_2 = 4$ and $w_1 = 0.3$. When $w$ is very low then the corresponding agents get poor results and they are no longer the winners, as in the case of very high $n$ in Figure 5.

---

8. This is in fact an illuminating instance of the well-known prisoners dilemma (Axelrod, 1984).

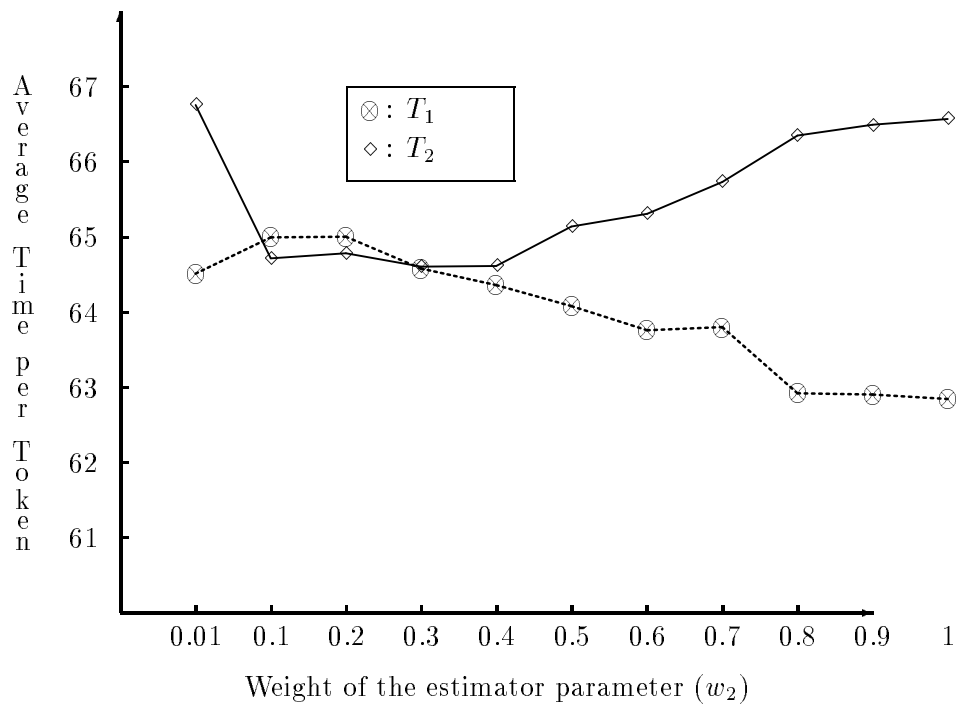

Figure 7: Performance of 2 populations of 50 agents with $n_1 = n_2 = 4$ and $w_1 = 0.3$

Another interesting phenomenon is obtained when confronting adaptive agents with *load-querying* agents. Load-querying agents are agents who are able to consult the resources about where they should submit their jobs. A load-querying agent will submit its job to the most unloaded resource at the given point. When confronting load-querying agents with adaptive ones, the results obtained by the adaptive agents are obviously worse than the results obtained by the load-querying ones, but are better than the results obtained by a complete population of adaptive agents. This means that load-querying agents do not play the role of "parasites", as the above-mentioned "exploiters"; the load-querying agents help in maintaining the load balancing among the resources, and therefore help the rest of the agents. Another result we obtain is that agents who adopt deterministic SRs may behave as parasites and worsen the performance of adaptive agents.

These assertions are supported by the experiments described in Figure 8, where a population of 90 agents, each of which uses an adaptive SR with parameters $(n, w)$, is faced with a minority of 10 agents which use different SRs, as stated above. In particular, in the four cases we consider, the minority behaves in the following ways: (*i*) they choose the resource which gave best results, (*ii*) they are very conservative in updating the history, (*iii*) they are load-querying agents, (*iiii*) they all use deterministically the resource with capacity 40 (in our basic experimental setting).

| 90 agents | 10 agents | $T_1$ | $T_2$ |
|-----------|-----------|-------|-------|
| (.3,4) | (.3,20) | 65.161 | 59.713 |
| (.3,4) | (.1,4) | 64.630 | 63.818 |
| (.3,4) | Load-querying | 62.320 | 47.236 |
| (.3,4) | Using Res. 0 | 65.499 | 55.818 |

Figure 8: Performance of 2 populations of 90/10 agents with various SRs

## 7. Communication among Agents

Up to this point, we have assumed that there is no direct communication among the agents. The motivation for this was that we considered situations in which there were absolutely no transmission channels and protocols. This assumption is in agreement with the idea of multi-agent reinforcement learning. In systems where massive communication is feasible we are not so much concerned with multiple agent adaptation, and the problem reduces to supplying satisfactory communication mechanisms. Multi-agent reinforcement learning is most interesting where real life forces agents to act without a-priori arranged communication channels and we must rely on action-feedback mechanisms. However, it is of interest to understand the effects of communication on the system efficiency (as in Shoham & Tennenholtz, 1992; Tan, 1993), where the agents are augmented with some sort of communication capabilities. Our study of this extension led to some illuminating results, which we will now present.

We assume that each agent can communicate only with some of the other agents, which we call its *neighbors*. We therefore consider a relation *neighbor-of* and assume it is reflexive, symmetric and transitive. As a consequence, the relation neighbor-of partitions the population into equivalence classes, that we call *neighborhoods*.

The form of communication we consider is based on the idea that the efficiency estimators of agents within a neighborhood will be shared among them when a decision is made (i.e., when an agent chooses a resource). The reader should notice that this is a naive form of communication and that more sophisticated types of communication are possible. However, the above form of communication is most natural when we concentrate on agents that update their behavior based only on past information. In particular, this type of communication is similar to the ones used in the above-mentioned work on incorporating communication into the framework of multi-agent reinforcement learning.

We suppose that different SRs may be used by different agents in the same population, but we impose the condition that within a single neighborhood, the same SR is used by all its members.

We also assume that each agent keeps its own history and updates it by itself in the usual way. The choice, instead, is based not only on the agent efficiency estimator, but on the average of the efficiency estimators of the agents in the corresponding neighborhood. Such average is called the *neighborhood efficiency estimator*. The neighborhood efficiency estimator has no physical storage: Its value is recalculated each time a member needs it.

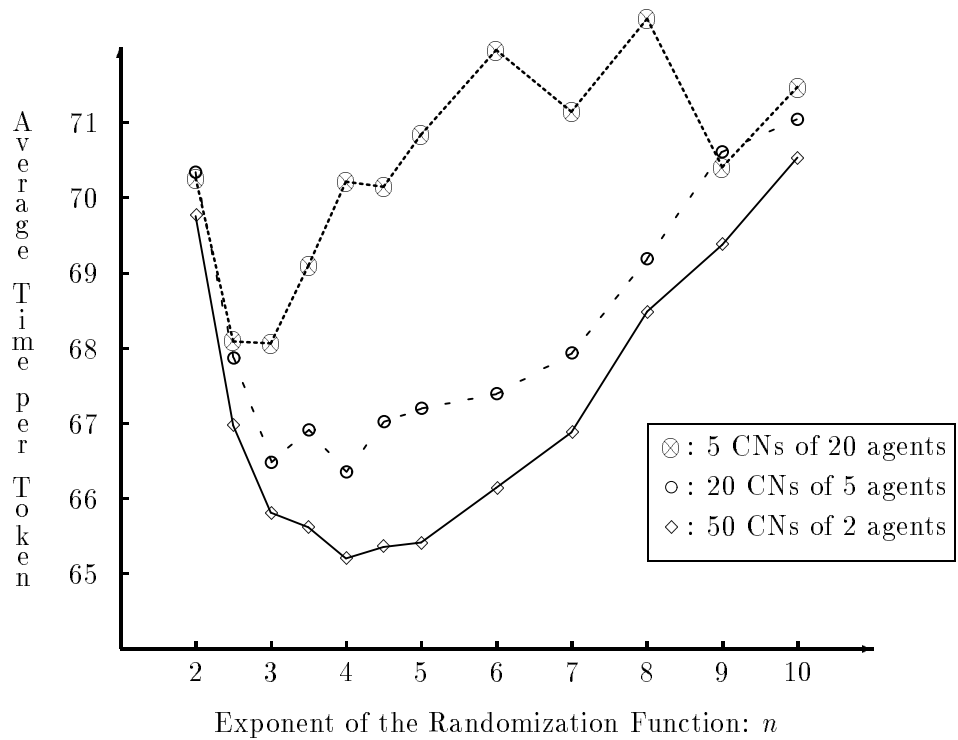

Figure 9: Performance of the adaptive Selection Rules for random load profile for communicating agents

In order to compare the behavior of communicating agents and non-communicating ones, we assume that in a single population there might be, aside from the neighborhoods defined above, also some neighborhoods that do not allow the sharing of efficiency estimators among its members. The members of these neighborhoods behave as described in the previous sections, i.e., each agent relies only on its own history. The only thing that is common among the members of such a neighborhood is that all its members use the same SR.

We call *communicating neighborhood* (CN), a neighborhood in which the efficiency estimators are shared when a decision is taken and *non-communicating neighborhood* (NCN), a neighborhood in which this is not done.

The first set of experiments we ran, regards a population composed of only CNs, all of the same size. In particular, we considered CNs of various sizes, starting from 50 CNs of size 2, going to 5 CNs of size 20. The load profile exploited is the random load change defined in Subsection 5.3, the value of $w$ is taken to be 0.3, and $n$ is taken to have various values. The results obtained are shown in Figure 9.

The results show that such communicating populations do not get good results. The reason for this is that members of a CN tend to be very conservative, in the sense that they mostly use the best resource. In fact, since they rely on an average of several agents, the picture they have of the system tends to be much more static. In particular, the bigger is the CN the more conservative its members tend to be. For example, consider the values of $(n, w)$ that give the best results for non-communicating agents, those values give quite bad performance for CNs since they turn to be too conservative.

Using more adaptive values of $(n, w)$, the behavior of a communicating population improves and reaches a performance that is just slightly worse than the performance of a non-communicating population. Tuning the parameters using a finer grain, it is possible to obtain a performance that is equal to the one obtained by a non-communicating population. However, it seems clear that no obvious gain is achieved from this form of communication capability. The intuitive explanation is that there are two opposite effects caused by the communication. On the one hand, the agents get a fairer picture of the system which prevents them from using bad resources and therefore getting bad performance. On the other hand, since all of the agents in a CN have a "better" picture of the system, they all tend to use the best resources and thus they all compete for them. In fact, the agents behave selfishly and their selfish interest may not agree with the interest of the population as a whole.

The interesting message that we get is that the fact that some agents may have a "distorted" picture of the system (which is typical for non-communicating populations), turns out to be an advantage for the population as a whole.

Sharing the data among agents leads to poorer performances also because in this case the agents have common views of loads and target jobs toward the same (lightly loaded) resources, which quickly become overloaded. In order to profitably use the shared data, we should allow for some form of reasoning about the fact that the data is shared. This problem however is out of the scope of this paper (see e.g., Lesser, 1991).

In order to understand the behavior of the system when CNs and NCNs face each other, we consider an NCN of 80 agents together with a set of CNs of equal size, for different values of that size. The results of the corresponding experiments are shown in Figure 10. The members of the CNs, being more inclined to use the best resources, behave as parasites in the sense explained in Section 6. They exploit the adaptiveness of the rest of the population to obtain good performance from the best resources. For this reason they get better results than the rest of the population, as shown by the experimental results.

It it interesting to observe that when the NCN uses a very conservative selection rule, the CNs obtain even better results. The intuitive explanation for this behavior is that although all groups, i.e., both the communicating ones and the one with high value of $n$, tend to be conservative, the communicating ones "win" because they are conservative in a more "clever" way, that is making use of a better picture of the situation.

The conclusion we draw in this section is that the proposed form of communication between agents may not provide useful means to improve the performance of a population in our setting. However, we do not claim that communication between agents is completely useless. Nevertheless, we have observed that it does not provide a straightforward significant improvement. Our results support the claim that the sole past history of an agent is a reasonable information on which to base its decision, assuming we do not consider available any kind of real-time information (e.g., current load of the resources).

| 80 agents | 20 agents | $T_1$ | $T_2$ |
|---|---|---|---|
| (.3,4) 1 NCN | (.3,4) 1 CN | 65.287 | 63.054 |
| (.3,4) 1 NCN | (.3,4) 2 CNs | 65.069 | 63.307 |
| (.3,4) 1 NCN | (.3,4) 5 CNs | 65.091 | 62.809 |
| (.3,4) 1 NCN | (.3,4) 10 CNs | 64.895 | 63.840 |
| (.3,10) 1 NCN | (.3,4) 1 CN | 68.419 | 60.018 |
| (.3,10) 1 NCN | (.3,4) 2 CNs | 68.319 | 59.512 |
| (.3,10) 1 NCN | (.3,4) 5 CNs | 68.529 | 60.674 |
| (.3,10) 1 NCN | (.3,4) 10 CNs | 68.351 | 61.711 |

Figure 10: Performance of CNs and NCNs together

## 8. Discussion

The previous sections were devoted to a report on our experimental study. We now synthesize our observations in view of our motivation, as discussed in Sections 1 and 2.

As we mentioned, our model is a general model where active autonomous agents have to select among several resources in a dynamic fashion and based on local information. The fact that the agents use only local information makes the possibility of efficient load-balancing questionable. However, we showed that adaptive load balancing based on purely local feedback is a feasible task. Hence, our results are complementary to the ones obtained in the distributed computer systems literature. As Mirchandaney and Stankovic (1986) put it: "...what is significant about our work is that we have illustrated that is possible to design a learning controller that is able to dynamically acquire relevant job scheduling information by a process of trial and error, and use that information to provide good performance." The study presented in our paper supplies a complementary contribution where we are able to show that useful adaptive load balancing can be obtained using *purely local information* and in the framework of a general organizational-theoretic model.

In our study we identified various parameters of the adaptive process and investigated how they affect the efficiency of adaptive load balancing. This part of our study supplies useful guidelines for a systems designer who may force all the agents to work based on a common selection rule. Our observations, although somewhat related to previous observations made in other contexts and models (Huberman & Hogg, 1988), enable to demonstrate aspects of purely local adaptive behavior in a non-trivial model.

Our results about the disagreement between selfish interest of agents and the common interest of the population is in sharp contrast to previous work on multi-agent learning (Shoham & Tennenholtz, 1992, 1994) and to the dynamic programming perspective of earlier work on distributed systems (Bertsekas & Tsitsiklis, 1989). Moreover, we explore how the interaction between different agent types affects the system's efficiency as well as the individual agent's efficiency. The related results can be also interpreted as guidelines for a designer who may have only partial control of a system.

The synthesis of the above observations teaches us about adaptive load balancing when one adopts a reinforcement learning perspective where the agents rely only on their local information and activity. An additional step we performed attempts to bridge some of the gap between our local view and previous work on adaptive load balancing by communicating agents, whose decisions may be controlled by learning automata or by other means. We therefore rule out the possibility of communication about the current status of resources and of joint decision-making, but enable a limited sharing of previous history. We show that such limited communication may not help, and even deteriorate system efficiency. This leaves us with a major gap between previous work where communication among agents is the basic tool for adaptive load balancing and our work. Much is left to be done in attempting to bridge this gap. We see this as a major challenge for further research.

## 9. Related Work

In Section 2 we mentioned some related work in the field of distributed computer systems (Mirchandaney & Stankovic, 1986; Billard & Pasquale, 1993; Glockner & Pasquale, 1993; Mirchandaney et al., 1989; Zhou, 1988; Eager et al., 1986). A typical example of such work is the paper by Mirchandaney and Stankovic (1986). In this work learning automata are used in order to decide on the action to be taken. However, the suggested algorithms heavily rely on communication and information sharing among agents. This is in sharp contrast to our work. In addition, there are differences between the type of model we use and the model presented in the above-mentioned work and in other work on distributed computer systems.

Applications of learning algorithms to load balancing problems are given by Mehra (1992), Mehra and Wah (1993). However, in that work as well, the agents (sites, in the authors' terminology) have the ability to communicate and to exchange workload values, even though such values are subject to uncertainty due to delays. In addition, differently from our work, the learning activity is done off-line. In particular, in the learning phase the whole system is dedicated to the acquisition of workload indices. Such load indices are then used in the running phase as threshold values for job migration between different sites.

In spite of the differences, there are some similarities between our work and the above-mentioned work. One important similarity is the use of learning procedures. This is in difference from the more classical work on parallel and distributed computation (Bertsekas & Tsitsiklis, 1989) which applies numerical and iterative methods to the solution of problems in network flow and parallel computing. Other similarities are related to our study of the division of the society into groups. This somewhat resembles work on group formation (Billard & Pasquale, 1993) in distributed computer systems. The information sharing we allow in Section 7 is similar to the limited communication discussed by Tan (1993). In the classification of load-balancing problems given by Ferrari (1985), our work falls into the category of *load-independent and non-preemptive pure load-balancing*. The problems we investigate can be also seen as *sender-initiated* problems, although in our case the sender is the agent and not the (overloaded) resource.

One may wonder how our work differs from other work on adaptive load balancing in Operations Research (OR) (e.g., queuing theory Bonomi, Doshi, Kaufmann, Lee, & Kumar, 1990). Indeed, there are some commonalities. In both OR and our work, individual decisions are made locally, based on information obtained dynamically during runtime. And in both cases the systems constructed are sufficiently complex that the most interesting results tend to be obtained experimentally. However, a careful look at the relevant OR literature reveals an essential difference between the perspective of OR on the topic and our reinforcement-learning perspective: OR permits free communication within the system, and thus there is no significant element of uncertainty in that framework. In particular, the issue of exploration versus exploitation, which lies at the heart of our approach, is completely absent from work in OR.

Some work on adaptive load balancing and related topics has been carried out also by the Artificial Intelligence community (see e.g., Kosoresow, 1993; Gmytrasiewicz, Durfee, & Wehe, 1991; Wellman, 1993). This work too, however, tends to be based on some form of communication among the agents, whereas in our case the load balancing is obtained purely from a learning activity.

This article is related to our previous work on *co-learning* (Shoham & Tennenholtz, 1992, 1994). The framework of co-learning is a framework for multi-agent learning, which differs from other frameworks discussed in multi-agent reinforcement learning (Narendra & Thathachar, 1989; Tan, 1993; Yanco & Stein, 1993; Sen, Sekaran, & Hale, 1994) due to the fact that it considers the case of *stochastic* interactions among *subsets* of the agents, where there is purely local feedback revealed to the agents based on these interactions. The framework of co-learning is similar in some respects to a number of dynamic frameworks in economics (Kandori, Mailath, & Rob, 1991), physics (Kinderman & Snell, 1980), computational ecologies (Huberman & Hogg, 1988), and biology (Altenberg & Feldman, 1987). Our study of adaptive load balancing can be treated as a study in co-learning.

Relevant to our work is also the literature in the field of *Learning Automata* (see Narendra & Thathachar, 1989). In fact, an agent in our setting can be seen as a learning automaton. Therefore, one may hope that theoretical results on interconnected automata and N-player games (see e.g., El-Fattah, 1980; Abdel-Fattah, 1983; Narendra & Wheeler Jr., 1983; Wheeler Jr. & Narendra, 1985) could be imported in our framework. Unfortunately, due to the stochastic nature of job submissions (i.e., agent interactions) and the real-valued (instead of binary) feedback, our problem does not fit completely in to the theoretical framework of learning automata. Hence, results concerning optimality, convergence or expediency of learning rules such as Linear Reward-Penalty or Linear Reward-Inaction, can not be easily adapted into our setting. The fact that we use a stochastic model for the interaction among agents, makes our work closely related to the above-mentioned work on co-learning. Nevertheless, our work is largely influenced by learning automata theory and our resource-selection rules closely resemble reinforcement schemes for learning automata.

Last but not least, our work is related to work applying organization theory and management techniques to the field of Distributed AI (Fox, 1981; Malone, 1987; Durfee, Lesser, & Corkill, 1987). Our model is closely related to models of decision-making in management and organization theory (e.g., Malone, 1987) and applies a reinforcement learning perspective to that context. This makes our work related to psychological models of decision-making (Arthur, 1994).

## 10. Summary

This work applies the idea of multi-agent reinforcement learning to the problem of load balancing in a loosely-coupled multi-agent system, in which agents need to adapt to one another as well as to a changing environment. We have demonstrated that adaptive behavior is useful for efficient load balancing in this context and identified a pair of parameters that affect that efficiency in a non-trivial fashion. Each parameter, holding the other parameter to be fixed, gives rise to a certain tradeoff, and the two parameters interplay in a non-trivial and illuminating way. We have also exposed illuminating results regarding heterogeneous populations, such as how a group of parasitic less adaptive agents can gain from the flexibility of other agents. In addition, we showed that naive use of communication may not improve, and might even deteriorate, the system efficiency.

## Acknowledgments

We thank the anonymous reviewers and Steve Minton, whose stimulating comments helped us in improving on an earlier version of this paper.